\pdfoutput=1

\documentclass[11pt]{article}

\usepackage[preprint]{acl}

\usepackage{times}
\usepackage{latexsym}
\usepackage{hyperref}
\usepackage{color, colortbl}
\definecolor{Gray}{gray}{0.9}
\usepackage{array}
\usepackage{booktabs}
\usepackage{multirow}
\usepackage{amsmath}
\usepackage{amssymb}
\usepackage{graphicx}
\usepackage{enumerate}
\usepackage{diagbox}
\usepackage{mathtools}
\usepackage{paralist}
\usepackage{tabularx}
\usepackage{lipsum}
\usepackage{ragged2e}

\usepackage[T1]{fontenc}

\usepackage[utf8]{inputenc}

\usepackage{microtype}

\usepackage{inconsolata}

\usepackage{graphicx}
\usepackage{tabularx}
\usepackage{booktabs}
\usepackage{rotating}
\usepackage{multirow}
\usepackage{setspace}
\usepackage{courier} 
\usepackage{url}            
\usepackage{booktabs}       
\usepackage{amsfonts}       
\usepackage{nicefrac}       
\usepackage{microtype}      
\usepackage{xcolor}         
\usepackage{amssymb}
\usepackage{pifont}
\usepackage[edges]{forest}

\definecolor{hidden-red}{RGB}{205, 44, 36}
\definecolor{hidden-blue}{RGB}{194,232,247}
\definecolor{hidden-orange}{RGB}{243,202,120}
\definecolor{hidden-green}{RGB}{34,139,34}
\definecolor{hidden-pink}{RGB}{255,245,247}
\definecolor{hidden-black}{RGB}{20,68,106}

\title{On the Role of Entity and Event Level Conceptualization in Generalizable Reasoning: A Survey of Tasks, Methods, Applications, and Future Directions}

\author{
\textbf{
Weiqi Wang,
Tianqing Fang,
Haochen Shi,
Baixuan Xu,
Wenxuan Ding,}\\
\textbf{
Liyu Zhang,
Wei Fan,
Jiaxin Bai,
Haoran Li,
Xin Liu,
Yangqiu Song}\\
Department of Computer Science and Engineering, HKUST, Hong Kong SAR, China\\
\texttt{\{wwangbw, tfangaa, yqsong\}@cse.ust.hk} \\ 
}

\begin{document}
\maketitle
\begin{abstract}
Conceptualization, a fundamental element of human cognition, plays a pivotal role in human generalizable reasoning.
Generally speaking, it refers to the process of sequentially abstracting specific instances into higher-level concepts and then forming abstract knowledge that can be applied in unfamiliar or novel situations. 
This enhances models' inferential capabilities and supports the effective transfer of knowledge across various domains.
Despite its significance, the broad nature of this term has led to inconsistencies in understanding conceptualization across various works, as there exists different types of instances that can be abstracted in a wide variety of ways.
There is also a lack of a systematic overview that comprehensively examines existing works on the definition, execution, and application of conceptualization to enhance reasoning tasks.
In this paper, we address these gaps by first proposing a categorization of different types of conceptualizations into four levels based on the types of instances being conceptualized, in order to clarify the term and define the scope of our work.
Then, we present the first comprehensive survey of over 150 papers, surveying various definitions, resources, methods, and downstream applications related to conceptualization into a unified taxonomy, with a focus on the entity and event levels.
Furthermore, we shed light on potential future directions in this field and hope to garner more attention from the community.
\end{abstract}

\section{Introduction}
\label{sec:introduction}

\begin{quote}
  ``\textit{Concepts are the glue that holds our mental world together.}''-- \citet{murphy2004big}
\end{quote}

\begin{figure}[t]
     \centering
     \includegraphics[width=1\linewidth]{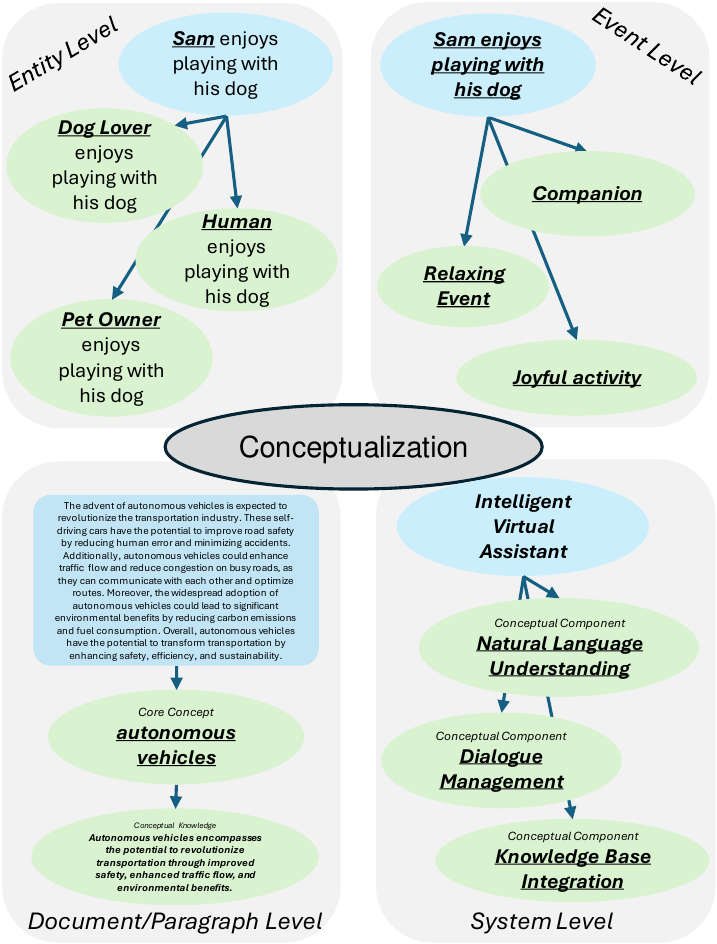}
     \caption{Examples of performing conceptualization at different semantic levels.}
    \label{fig:introduction_examples}
    \vspace{-0.1in}
\end{figure}

Conceptualization has been widely recognized as a fundamental component of human intelligence, spanning fields from psychology~\cite{kahneman2011thinking,evans2003two,bransford1971abstraction} to computational linguistics~\cite{DBLP:journals/cacm/BengioLH21,tenenbaum2011grow,DBLP:journals/tacl/LachmyPMT22}.
In the era of deep learning, numerous studies have emerged focusing on conceptualization as a means to achieve generalizable reasoning with (Large) Language Models (LLMs;~\citealp{openai2022chatgpt,GPT4,llama,llama2,Gemma,Gemini}) in areas such as commonsense reasoning~\cite{CAT,CAR,CANDLE}, causal reasoning~\cite{DBLP:journals/coling/FederOSR21,DBLP:journals/cogsci/KundaMC90}, physical reasoning~\cite{DBLP:conf/aaai/BiskZLGC20,DBLP:conf/emnlp/WangDFS23,DBLP:conf/nips/HongYTTG21}, and more.

In general terms, conceptualization refers to the process of consolidating specific instances with shared properties or characteristics into a cohesive concept that represents a vast collection of instances. 
It is a sub-type of abstraction~\cite{DBLP:journals/ai/GiunchigliaW92}, but specifically requires the presence of a concept as the base for such abstraction.
With proper conceptualization, abstract knowledge can be subsequently derived by associating original knowledge at the instance level with that concept.
When encountering unfamiliar or novel scenarios, concepts in abstract knowledge can be instantiated to new instances to support downstream reasoning~\cite{tenenbaum2011grow}.
This process can occur at various levels, including entity~\cite{Probase,Probase+,DBLP:conf/emnlp/AlukaevKPIIKT23,DBLP:conf/acl/LiuYWWPZS23,DBLP:conf/acl/00050FS25}, event~\cite{AbstractATOMIC,CANDLE,AbsPyramid}, paragraph/document~\cite{DBLP:conf/naacl/FalkeG19,DBLP:conf/ijcnlp/FalkeMG17}, and system levels~\cite{ConceptualizationNLPTasks,DBLP:conf/aaai/KadiogluK24}, ultimately forming a hierarchy that contribute to a comprehensive understanding and representation of knowledge.

Despite its significance, the field lacks a comprehensive and unified taxonomy to categorize existing research on conceptualization. 
On the one hand, the term ``conceptualization'' is inherently broad, encompassing various types of conceptualizations across different instances and performed in various ways, all included under a single term.
As illustrated in Figure~\ref{fig:introduction_examples}, the conceptualization of entities and documents requires two distinct paradigms; however, the current terminology fails to adequately address these differences.
This has led to confusion and miscommunication among works that apply conceptualization in their methodologies.
On the other hand, the methods for conceptualizing different types of instances in a scalable and accurate manner remain unclear.
Finally, it is essential to summarize the benefits that conceptualization can bring to downstream tasks to gather insights for future applications and new research directions.

To address these issues, we present the first-ever survey that systematically taxonomizes conceptualization. 
First, in Section~\ref{sec:definitions}, we define four types of conceptualization based on different semantic levels of the instances being conceptualized: entity, event, document, and system. 
In later sections, we focus on two main types of conceptualization based on the entity and event levels, as they are largely uncharted in existing literature and play a key role in human reasoning. 
We then propose a set of objectives to select and survey papers that feature conceptualization as their core idea, review more than 150 papers, and organize them into three main categories, as shown in Figure~\ref{fig:taxonomy}.
We summarize the main representative tasks and datasets available for these types of conceptualization in Section~\ref{sec:tasks_and_datasets}.  
Subsequently, in Section~\ref{sec:methods}, we categorize conceptualization acquisition methods into extraction, retrieval, and generative-based methods. 
The downstream benefits of conceptualization are discussed in Section~\ref{sec:downstream_applications}, with a specific focus on several reasoning tasks. 
Finally, in Section~\ref{sec:future-directions}, we propose two future directions that can benefit from conceptualization. 
We hope our work can serve as a practical handbook for researchers and pave the way for further advancements in the field of conceptualization.


\section{Four Levels of Conceptualization}
\label{sec:definitions}
We first define four levels of conceptualization according to the type of instances being conceptualized. 
They are categorized into four levels: entity level, event level, document level, and system level.
Running examples are shown in Figure~\ref{fig:introduction_examples}.

\tikzstyle{my-box}=[
    rectangle,
    draw=hidden-black,
    rounded corners,
    text opacity=1,
    minimum height=1.5em,
    minimum width=5em,
    inner sep=2pt,
    align=center,
    fill opacity=.5,
]
\tikzstyle{leaf}=[
    my-box, 
    minimum height=1.5em,
    fill=hidden-blue!90, 
    text=black,
    align=left,
    font=\normalsize,
    inner xsep=2pt,
    inner ysep=4pt,
]
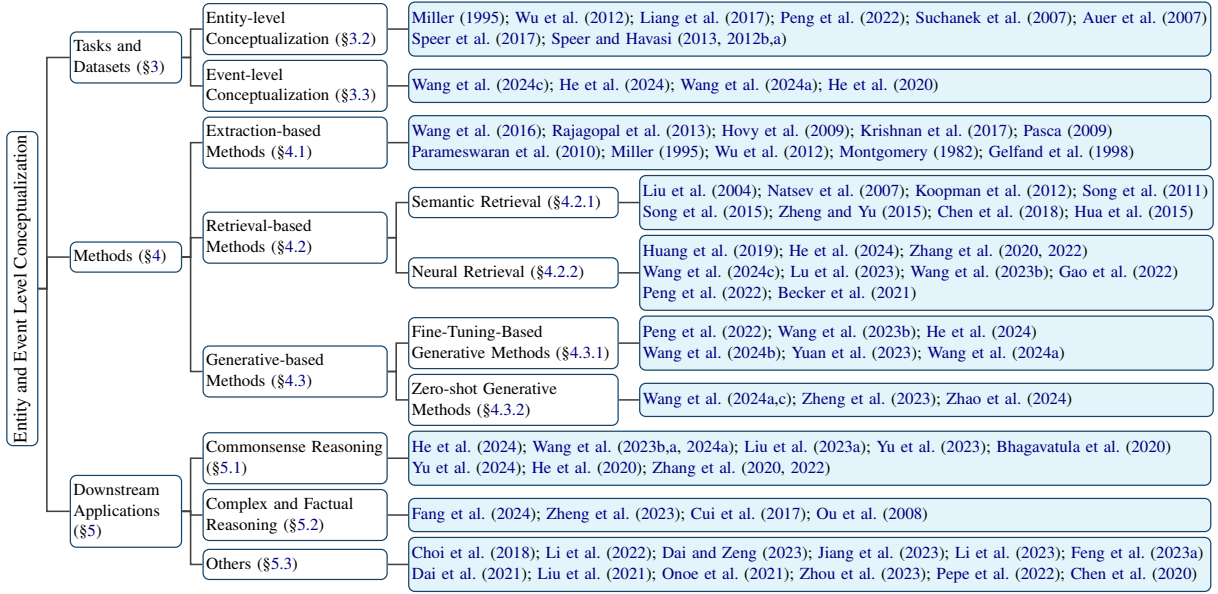
\begin{figure*}[t]
    \vspace{-2mm}
    \centering
    \resizebox{1\textwidth}{!}{
        \begin{forest}
            forked edges,
            for tree={
                grow=east,
                reversed=true,
                anchor=base west,
                parent anchor=east,
                child anchor=west,
                base=left,
                font=\large,
                rectangle,
                draw=hidden-black,
                rounded corners,
                align=left,
                minimum width=4em,
                edge+={darkgray, line width=1pt},
                s sep=3pt,
                inner xsep=2pt,
                inner ysep=3pt,
                line width=0.8pt,
                ver/.style={rotate=90, child anchor=north, parent anchor=south, anchor=center},
            },
            where level=1{text width=6.2em,font=\normalsize,}{},
            where level=2{text width=10.5em,font=\normalsize,}{},
            where level=3{text width=12em,font=\normalsize,}{},
            where level=4{text width=12em,font=\normalsize,}{},
            [
                Entity and Event Level Conceptualization, ver
                [
                    Tasks and \\ Datasets (\S \ref{sec:tasks_and_datasets})
                    [
                        Entity-level \\ Conceptualization (\S \ref{sec:entity-level-abstraction})
                        [   
                            \citet{DBLP:journals/cacm/Miller95,Probase,Probase+,COPEN,DBLP:conf/www/SuchanekKW07,DBLP:conf/semweb/AuerBKLCI07}\\
                            \citet{DBLP:conf/aaai/SpeerCH17,DBLP:series/tanlp/SpeerH13,DBLP:conf/lrec/SpeerH12,DBLP:journals/tinytocs/SpeerH12}
                                , leaf, text width=47em
                        ]
                    ]
                    [
                        Event-level \\ Conceptualization (\S \ref{sec:event-level-abstraction})
                        [   
                            \citet{AbsPyramid,AbstractATOMIC,CANDLE,DBLP:journals/corr/abs-2003-03239}
                            , leaf, text width=47em
                        ]
                    ]
                ]
                [
                    Methods (\S \ref{sec:methods})
                    [
                        Extraction-based \\ Methods (\S \ref{sec:methods-extraction-based-methods})
                        [
                           \citet{DBLP:conf/cikm/WangOWWLPG16,DBLP:conf/www/RajagopalCOK13,DBLP:conf/emnlp/HovyKR09,DBLP:conf/cikm/KrishnanSZ017,DBLP:conf/eacl/Pasca09}\\
                           \citet{DBLP:journals/pvldb/ParameswaranGR10,DBLP:journals/cacm/Miller95,Probase,montgomery1982concept,gelfand1998automated}
                            , leaf, text width=47em
                        ]
                    ]
                    [
                        Retrieval-based \\ Methods (\S \ref{sec:methods-retrieval-based-methods})
                        [
                            Semantic Retrieval (\S \ref{sec:methods-retrieval-based-semantic-based-retrieval})
                            [
                                \citet{DBLP:conf/sigir/LiuLYM04,DBLP:conf/mm/NatsevHTXY07,DBLP:conf/cikm/KoopmanZBSL12,STC}\\\citet{ODSTC,DBLP:conf/emnlp/ZhengY15,DBLP:conf/cikm/ChenLXX18,DBLP:conf/icde/HuaWWZZ15}\\
                                , leaf, text width=33.5em
                            ]
                        ]
                        [
                            Neural Retrieval (\S \ref{sec:methods-retrieval-based-neural-based-retrieval})
                            [
                                \citet{DBLP:conf/emnlp/HuangSQH19,AbstractATOMIC,DBLP:conf/www/ZhangLPSL20,DBLP:journals/ai/ZhangLPKOFS22}\\\citet{AbsPyramid,CourseConceptExtraction,CAT,DBLP:conf/emnlp/GaoHKWMB22}\\\citet{COPEN,DBLP:conf/eacl/BeckerKF21}
                                , leaf, text width=33.5em
                            ]
                        ]
                    ]
                    [
                        Generative-based \\ Methods (\S \ref{sec:methods-generative-based-methods})
                        [
                            Fine-Tuning-Based\\Generative Methods (\S \ref{sec:methods-generative-based-fine-tuning-based-generative-methods})
                            [
                                \citet{COPEN, CAT, AbstractATOMIC}\\\citet{AbsInstruct,CCE,CANDLE}
                                , leaf, text width=33.5em
                            ]
                        ]
                        [
                            Zero-shot Generative \\ Methods (\S \ref{sec:methods-generative-based-zero-shot-generative-methods})
                            [
                                \citet{CANDLE,AbsPyramid,StepBackPrompt,HIConcept} 
                                , leaf, text width=33.5em
                            ]
                        ]
                    ]
                ]
                [
                    Downstream \\ Applications \\ (\S \ref{sec:downstream_applications})
                        [
                            Commonsense Reasoning \\ (\S \ref{section:commonsense_reasoning})
                            [
                                \citet{AbstractATOMIC,CAT,CAR,CANDLE,DimonGen,DBLP:conf/acl/YuWLBSLG0Y23,DBLP:conf/iclr/BhagavatulaBMSH20}\\
                                \citet{yu2024cosmo,DBLP:journals/corr/abs-2003-03239,DBLP:conf/www/ZhangLPSL20,DBLP:journals/ai/ZhangLPKOFS22}
                                , leaf, text width=47em
                            ]
                        ]
                        [
                            Complex and Factual \\ Reasoning (\S \ref{sec:downstream-reasoning-complex})
                            [
                                \citet{DBLP:journals/corr/abs-2403-07398,StepBackPrompt,DBLP:journals/pvldb/CuiXWSHW17,DBLP:conf/lrec/OuPOSN08,DBLP:conf/emnlp/0001WKLFBLYLLYY24,DBLP:conf/emnlp/Xu0S0JFBLYLLYYC24}
                                , leaf, text width=47em
                            ]
                        ]
                        [
                            Others (\S \ref{sec:downstream-reasoning-others})
                            [
                                \citet{DBLP:conf/acl/LevyZCC18,DBLP:journals/tacl/LiYC22,DBLP:conf/acl/DaiZ23,DBLP:conf/acl/JiangHJWXT23,DBLP:conf/emnlp/LiBS23,DBLP:conf/emnlp/FengPM23}\\
                                \citet{DBLP:conf/acl/DaiSW20,DBLP:conf/emnlp/LiuLXH0W21,DBLP:conf/acl/OnoeBMD20,DBLP:conf/aaai/Zhou00023,DBLP:conf/aaai/PepeBBN22,DBLP:conf/conll/ChenZWR20}
                                , leaf, text width=47em
                            ]
                        ]
                ]
            ]
        \end{forest}
    }
    \vspace{-0.1in}
    \caption{Taxonomy of representative works in entity and event level \textbf{conceptualization} categorized by tasks and datasets (\S \ref{sec:tasks_and_datasets}), methods in performing conceptualization (\S \ref{sec:methods}), and downstream applications (\S \ref{sec:downstream_applications}).}
    \label{fig:taxonomy}
    \vspace{-0.2in}
\end{figure*}

\noindent\textbf{Entity Level:}
Entity-level conceptualization involves grouping multiple entities under a shared concept~\cite{DBLP:conf/acl/YangZNZP20,COPEN}.  
It is the most common form of conceptualization in human cognition and is frequently applied for knowledge acquisition~\cite{carey1991knowledge,murphy2004big}. 
For instance, entities like ``apple,'' ``pear,'' and ``grape,'' can be categorized together under the broader concept of ``fruit.''  
By doing so, abstract knowledge can be derived by reintegrating the concept into the context of specific instances, such as the assertion ``fruit is delicious,'' with ``apple is delicious'' serving as the specific source. 
When someone encounters an unknown fruit, they can quickly understand its properties by associating it with the abstract knowledge of fruit, such as its possible taste or nutrition.

\noindent\textbf{Event Level:}
While a concept can capture the semantic meaning of a group of entities, it can also represent events at a higher level of conceptualization. 
Event-level conceptualization aims to broaden the scope from entities to include events as well~\cite{AbstractATOMIC,AbsPyramid}. 
It seeks to associate different events under a shared concept that preserves the original semantic meaning to the maximum extent possible. 
For instance, activities like ``Sam playing with his dog,'' ``Alex dancing in the club,'' and ``Bob doing yoga'' can all be conceptualized as ``relaxing events.'' 
Abstract knowledge can then follow, stating that ``If someone engages in relaxing events, they feel happy and relaxed.''  
When someone encounters an unknown or unfamiliar event, such as ``Charlie likes painting the sunset,'' they can infer that painting the sunset is a relaxing event and that Charlie feels happy and relaxed when doing so.

\noindent\textbf{Document Level:}
Document-level conceptualization further extends the scope of the instance from entities and events to paragraphs or even entire documents. 
It aims to generate a summary that captures the main ideas and essential information while maintaining the overall semantic and context of the original text.
Previous works on abstractive summarization~\cite{DBLP:conf/acl/LadhakD0CM22,DBLP:conf/emnlp/WangGHZ19} have identical objectives, and earlier surveys by~\citet{DBLP:journals/tacl/RennardSHV23,DBLP:conf/aaai/LinN19,DBLP:conf/acl/LiuLYZJLH24} have effectively summarized these studies. 
Therefore, we only mention it here to clarify document-level conceptualization for readers and will not go into further detail in later sections to avoid overlap.

\noindent\textbf{System Level:}
Finally, system-level conceptualization aims to simplify the understanding of a complex system by abstracting its behavior and functionality into a higher-level representation. 
It is derived from the design of operating systems~\cite{DBLP:journals/hhci/DoanePK90} and is under-studied in the domain of NLP. 
The only representative example is a recent work by~\citet{ConceptualizationNLPTasks}, where the authors provide a systematic categorization of NLP tasks based on their objectives and characteristics while neglecting the detailed format of input/output and the datasets on which the tasks are evaluated. 
Due to the limited number of works available, we will not survey this type of conceptualization.

In later sections, we focus specifically on entity and event-level conceptualizations and propose a taxonomy to categorize works into three categories. 
To ensure that our search for papers is comprehensive and objective in relation to our target scope, we propose the following three objectives for selecting the most relevant papers.
First, we aim for papers that adhere to the paradigm of linking different instances together and use concepts as representations of the formed clusters. 
We also seek papers that aim to establish hierarchies between different entities and events. 
Finally, we look for papers that directly seek abstractions of entities or events via concepts.
Our proposed taxonomy is primarily categorized into resources, methods, and downstream applications of conceptualization, as this is the most straightforward structure for readers to grasp the topic.

\section{Tasks and Datasets}
\label{sec:tasks_and_datasets}
We first survey available datasets and benchmarks, as well as their associated tasks, for these two types of conceptualizations.
Statistical comparisons between different resources are shown in Table~\ref{tab:dataset_statistics}.
For datasets that also serve as evaluation benchmarks, we mark their associated tasks with classification task (\texttt{CLS}) and generation task (\texttt{GEN}).

\subsection{Concept Linking Task}
Most conceptualizations can be formulated as a concept linking task, where the goal is to link an instance $i$ to a concept $c$ such that $i$ can be semantically represented by $c$. 
It is challenging due to the infinite number of possible instance-concept pairs. 
Previous approaches, such as those by~\citet{DBLP:conf/www/BrauerHHLNB10,DBLP:conf/aaai/YatesGF15}, have attempted to further restrict the task to linking instances to a limited set of strict ontologies using heuristic or statistical methods. 
The task can also be formulated with a generative objective, which requires a model to generate $c$ directly given $i$ as input.

\subsection{Entity-level Conceptualization Datasets}
\label{sec:entity-level-abstraction}
To conceptualize different entities into concepts, multiple large-scale concept taxnomies have been constructed as resources for this type of conceptualization.
WordNet~\cite{DBLP:journals/cacm/Miller95} is the first and most well-known concept taxonomy, which is a large lexical database of English. 
It is a network of concepts, where each concept is a set of synonyms.
Probase~\cite{Probase,Probase+} is a later built concept taxonomy, which is a large-scale probabilistic taxonomy of concepts. 
It is constructed by analyzing a large amount of web pages and search logs.
YAGO~\cite{DBLP:conf/www/SuchanekKW07} is a semantic knowledge base, which is a large-scale concept taxonomy of entities and events. It is constructed by extracting information from Wikipedia~\cite{DBLP:conf/iclr/MerityX0S17} and WordNet.
DBPedia~\cite{DBLP:conf/semweb/AuerBKLCI07} is a large-scale knowledge base which is built by extracting structured information from Wikipedia. It also contains structured conceptual knowledge about entities and events.
ConceptNet~\cite{DBLP:conf/aaai/SpeerCH17} is the most recent concept taxonomy, featuring a large-scale semantic network of concepts. It is constructed by extracting structured information from various sources, including Wikipedia, WordNet, and Open Mind Common Sense~\cite{DBLP:conf/coopis/SinghLMLPZ02}.
Recently,~\citet{COPEN} introduced COPEN, a entity level conceptualization benchmark that is constructed by probing language models to retrieve concepts of an entity from a pre-defined set of concepts.
All of them are important knowledge bases that are rich in entity conceptualizations.

\begin{table}[t]
\centering
\resizebox{\linewidth}{!}{
\begin{tabular}{@{}l|l|ll|l@{}}
\toprule
Type & Dataset & \#Instance & \#Concept. & Tasks \\ 
\midrule
\multirow{7}{*}{\textbf{\textit{Entity}}} & WordNet & 82,115 & 84,428 & N/A \\
& Probase & 10,378,743 & 16,285,393 & N/A \\
& Probase+ & 10,378,743 & 21,332,357 & N/A \\
& YAGO & 143,210 & 352,297 & N/A \\ 
& DBPedia & 1,000,000 & 1,000,000 & N/A\\
& ConceptNet & 21,000,000 & 8,000,000 & N/A \\
& COPEN & 24,000 & 393 & \texttt{CLS} \\
\midrule
\multirow{3}{*}{\textbf{\textit{Event}}} & Abs.ATM. & 21,493 & 503,588 & \texttt{CLS,GEN} \\
& Abs.Pyr. & 17,000 & 220,797 & \texttt{CLS,GEN} \\
& CANDLE & 21,442 & 6,181,391 & N/A \\
\bottomrule
\end{tabular}
}
\vspace{-0.1in}
\caption{Statistical comparisons between different datasets with entity and event level conceptualizations.}
\vspace{-0.2in}
\label{tab:dataset_statistics}
\end{table}

\subsection{Event-level Conceptualization Datasets}
\label{sec:event-level-abstraction}
Compared to abstracting entities, there are fewer resources available for event-level conceptualizations. 
The most notable is the AbstractATOMIC dataset~\cite{AbstractATOMIC}, which was constructed by filtering head events from the ATOMIC dataset and identifying instance candidates within each event using syntactic parsing and human-defined rules. 
These instances are matched against Probase and WordNet to acquire candidate concepts using GlossBERT~\cite{DBLP:conf/emnlp/HuangSQH19}, which are then verified by a supervised model and human annotations. 
AbsPyramid~\cite{AbsPyramid} extends the AbstractATOMIC pipeline to ASER~\cite{DBLP:conf/www/ZhangLPSL20,DBLP:journals/ai/ZhangLPKOFS22}, a large-scale eventuality knowledge graph, by incorporating candidate concepts generated by ChatGPT to complement Probase and WordNet. 
It also extends coverage to verbs in addition to nouns and events, and broadens the domain of events from social aspects to all aspects. 
Both datasets provide rich event conceptualizations sourced from diverse origins.

\section{Conceptualization Acquisition Methods}
\label{sec:methods}
Next, we survey methods for performing or collecting entity and event-level conceptualizations.
We categorize them into three paradigms: extraction, retrieval, and generative-based methods, which are briefly demonstrated in Figure~\ref{fig:method_comparisons}.
We provide more discussions in Appendix~\ref{appendix:methods}.

\begin{figure*}[t]
     \centering
     \includegraphics[width=1\linewidth]{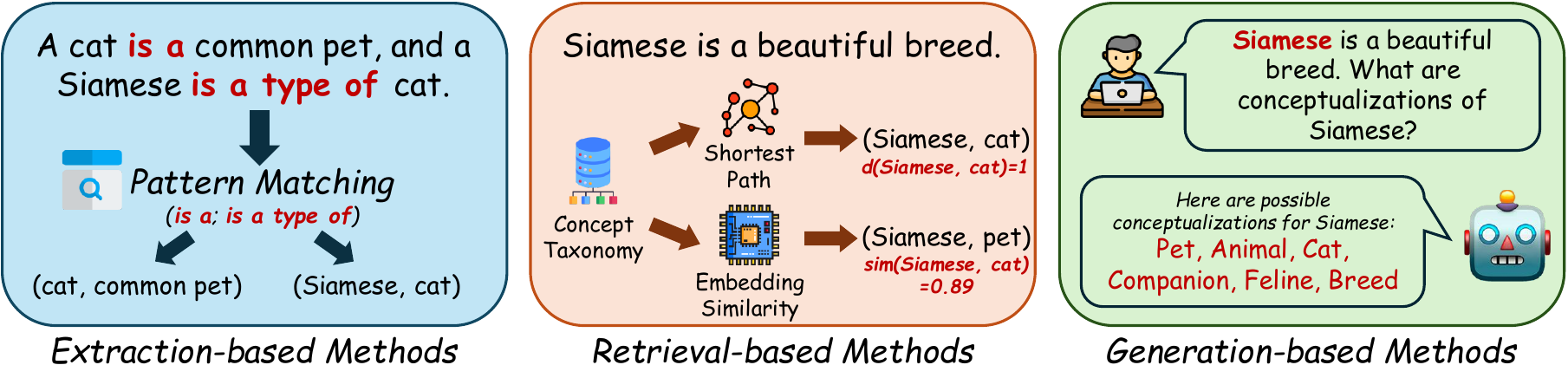}
     \vspace{-0.1in}
     \caption{Conceptual demonstration of different types of methods in performing or collecting entity and event level conceptualizations. 
     Instance and conceptualization pairs can be obtained at the end of each type of method.}
    \label{fig:method_comparisons}
    \vspace{-0.1in}
\end{figure*}

\subsection{Extraction-Based Methods}
\label{sec:methods-extraction-based-methods}
Extracting concepts from text is the earliest paradigm for systematically collecting conceptualizations~\cite{montgomery1982concept,gelfand1998automated}.
It typically involves first extracting all possible concepts from the text, followed with identifying the relationships between these concepts.
In this process, concepts are recognized either by looking for the most frequent words or by matching against a predefined list of patterns, such as ``is a,'', ``is a type of'', etc. 
Instances are then matched by looking for the subject of these patterns in the text, which forms instance-conceptualization pairs.
The main advantages of extraction-based methods~\cite{DBLP:conf/cikm/WangOWWLPG16,DBLP:journals/pvldb/ParameswaranGR10,DBLP:conf/www/RajagopalCOK13,DBLP:conf/emnlp/HovyKR09,DBLP:conf/cikm/KrishnanSZ017,DBLP:conf/eacl/Pasca09} are easy implementation, high processing speed, and free of training data.
This has facilitated the development of many large-scale concept taxonomies and knowledge bases, such as WordNet~\cite{DBLP:journals/cacm/Miller95}, ConceptNet~\cite{DBLP:conf/aaai/SpeerCH17,DBLP:series/tanlp/SpeerH13,DBLP:conf/lrec/SpeerH12,DBLP:journals/tinytocs/SpeerH12}, Probase~\cite{Probase,Probase+}, and DBpedia~\cite{DBLP:conf/semweb/AuerBKLCI07,DBLP:journals/ws/BizerLKABCH09}.
However, these methods, while successful in extracting conceptual relationships from text, are limited by text quality, reliance on predefined concepts, lack of semantic understanding, difficulty handling ambiguous words, and poor generalization to new domains or unseen concepts.

\subsection{Retrieval-Based Methods}
\label{sec:methods-retrieval-based-methods}
\subsubsection{Semantic-Based Retrieval}
\label{sec:methods-retrieval-based-semantic-based-retrieval}
Semantic-based retrieval methods aim to obtain conceptualizations by looking at the semantic similarity between the input instance and the concepts in a pre-defined concept taxonomy. 
It typically involves representing both the instance and a set of concepts into a shared semantic space and calculating the similarity between them.
One representative approach is to use WordNet~\cite{DBLP:journals/cacm/Miller95}, a large lexical database of English words, to calculate semantic similarity between two words as their shortest path in the WordNet hierarchy~\cite{DBLP:conf/sigir/LiuLYM04}.
Other methods~\cite{DBLP:conf/mm/NatsevHTXY07,STC,ODSTC,DBLP:conf/cikm/KoopmanZBSL12,DBLP:conf/emnlp/ZhengY15,DBLP:conf/cikm/ChenLXX18,DBLP:conf/icde/HuaWWZZ15} also share similar aspirations and define their own way of calculating such similarities.
However, these methods are usually limited by the need for comprehensive and accurate knowledge bases, high computational costs, the inability to handle unseen concepts, and the loss of important semantic context, prompting the development of neural-based retrieval methods.

\subsubsection{Neural-Based Retrieval}
\label{sec:methods-retrieval-based-neural-based-retrieval}
Neural-based retrieval methods overcome previous limitations by leveraging neural networks (or language models) to learn the semantic representations of the input instance and the concepts in the knowledge base or concept taxonomy. 
Then, the similarity between the input instance and the concepts can be calculated based on the learned representation embeddings.
This approach can be benefitted by the advancement in language modeling, such as BERT~\cite{DBLP:conf/naacl/DevlinCLT19}, RoBERTa~\cite{DBLP:journals/corr/abs-1907-11692}, and DeBERTa~\cite{DBLP:conf/iclr/HeLGC21,DBLP:conf/iclr/HeGC23}.
The most representative work in neural-based concept retrieval is AbstractATOMIC~\cite{AbstractATOMIC}. 
It uses GlossBERT~\cite{DBLP:conf/emnlp/HuangSQH19} to encode concepts (from WordNet and Probase) and instances (extracted from events in ATOMIC~\cite{DBLP:conf/emnlp/SapRCBC19}) into embeddings and leverage cosine similarity and human annotations to collect conceptualizations in a large scale manner.
Other methods~\cite{AbsPyramid,DBLP:conf/www/ZhangLPSL20,DBLP:journals/ai/ZhangLPKOFS22,CourseConceptExtraction,CAT,DBLP:conf/emnlp/GaoHKWMB22,DBLP:conf/eacl/BeckerKF21} similarly adopt different strategies in leveraging LMs as encoders, expanding the coverage of instances,training retrieval models.
Despite their promising results, these methods are limited by their need for extensive labeled data, reliance on the completeness and accuracy of the knowledge base, and inability to retrieve new concepts that are out of training data.

\subsection{Generative-Based Methods}
\label{sec:methods-generative-based-methods}
\subsubsection{Fine-Tuning-Based Generative Methods}
\label{sec:methods-generative-based-fine-tuning-based-generative-methods}
Fine-tuning-based generative methods aim to take an entity or event as input and generate the concept directly via a fine-tuned generative language model. 
This approach allows the model to generate conceptualizations for new instances and offers maximum flexibility of the input. 
Several methods~\cite{COPEN, CCE, AbstractATOMIC,AbsPyramid,AbsInstruct,CAT} have adopted this paradigm in training generative conceptualizers, based on models such as GPT2~\cite{radford2019language}, BART~\cite{DBLP:conf/acl/LewisLGGMLSZ20}, and T5~\cite{DBLP:journals/jmlr/RaffelSRLNMZLL20}, for automated conceptualization acquisition.
These methods typically train LMs on human-annotated or pre-existing conceptualization resources and yield outstanding results.
However, fine-tuning-based generative methods are limited by their high computational cost, time-consuming and resource-intensive data collection, uncertain performance across diverse domains, and relatively low quality of novel concepts compared to human annotations~\cite{liu-etal-2025-revisiting}. 
While these are common limitations associated with fine-tuned generative models, zero-shot generative methods using powerful LLMs and advanced prompting techniques potentially address these issues.

\subsubsection{Zero-Shot Generative Methods}
\label{sec:methods-generative-based-zero-shot-generative-methods}
Finally, zero-shot generative-based methods leverage powerful LLMs~\cite{DBLP:conf/nips/BrownMRSKDNSSAA20,openai2022chatgpt,GPT4,Gemini,llama,llama2} to generate the concept directly from an input instance. 
They rely on the vast amount of internal knowledge within the model and human-crafted prompts to efficiently distill conceptualizations and abstract knowledge from the models.
This is particularly useful when training data is scarce or when the domain is new and there are no existing training data available. 
Existing methods~\cite{CANDLE,AbsPyramid,StepBackPrompt,HIConcept} all share similar aspirations in collecting conceptualizations. 
The benefits are significant, as these methods can collect conceptualizations efficiently and at low cost without specific fine-tuning.
The resulting conceptualization knowledge base are thus scalable and downstream models trained on them typically have improved generalization ability to new instances and domains. 
However, to ensure high-quality generated conceptualizations, it is recommended to implement quality control mechanisms such as human evaluation or discriminators as post-filters. 
Recent studies~\cite{CANDLE,DBLP:journals/corr/abs-2403-07398} have shown that commonsense plausibility estimators~\cite{DBLP:conf/emnlp/0010WWS0H23} are effective for such quality control.

\section{Downstream Applications}
\label{sec:downstream_applications}
\begin{figure*}[t]
     \centering
     \includegraphics[width=1\linewidth]{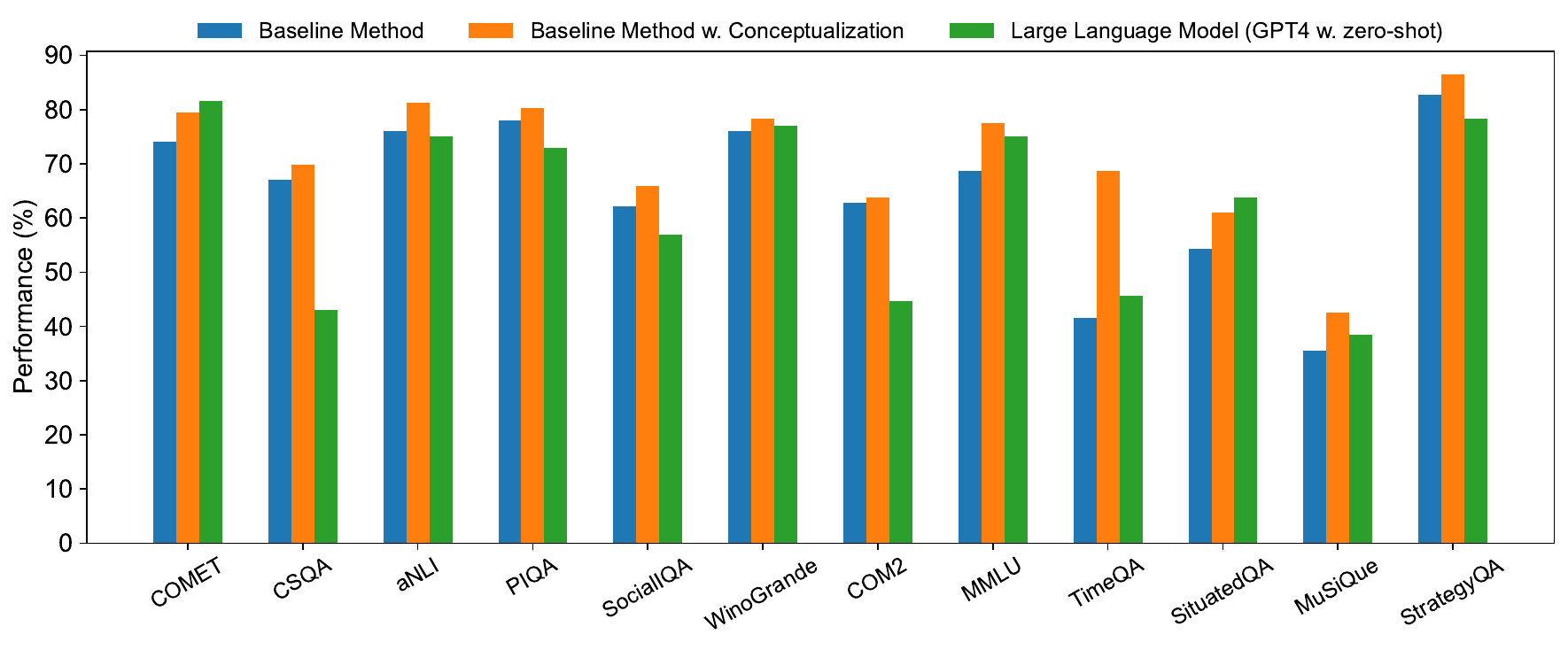}
     \vspace{-0.2in}
     \caption{Empirical benefits of conceptualization in methods across various benchmarks compared to baselines.}
    \label{fig:downstream}
    \vspace{-0.1in}
\end{figure*}

We then survey downstream tasks that can benefit from applying conceptualizations to provide readers with a general picture of what can be achieved and how to benefit from integrating conceptualizations.
An overview of performances by different methods that leverage conceptualization, evaluated on various benchmarks, are shown in Figure~\ref{fig:downstream}.

\subsection{Commonsense Reasoning}
\label{section:commonsense_reasoning}
Commonsense reasoning is the ability to make inferences about the world based on common knowledge, which involves reasoning about everyday events and situations~\cite{DBLP:books/daglib/0066824,DBLP:journals/cacm/DavisM15}. 
In this section, we discuss how conceptualizations benefit models in performing commonsense reasoning tasks.

\paragraph{Generative Commonsense Inference Modeling:}
The task of generative commonsense inference modeling (COMET;~\cite{DBLP:conf/acl/BosselutRSMCC19,DBLP:conf/aaai/HwangBBDSBC21}) aims to complete an inferential commonsense knowledge given a head event and a commonsense relation. 
State-of-the-art methods for COMET mainly fine-tune language models on large-scale commonsense knowledge bases, which suffer from data sparsity and lack of diversity in commonsense knowledge.
Although transfer from LLMs helps~\cite{DBLP:conf/naacl/WestBHHJBLWC22,DBLP:conf/emnlp/West0SLJLCHBB023}, distilled knowledge tends to be too easy for models to learn and converge to trivial inferences.
To address these issues, \citet{CAT} proposed to leverage conceptualization as knowledge augmentation tools to improve COMET.
Conceptualizations are first derived from head events to obtain abstracted events. 
Then, the tail of the original commonsense knowledge is placed back to the abstracted event to form abstracted commonsense knowledge. 
These derived abstract knowledge are then integrated with the original knowledge in commonsense knowledge bases to enrich the diversity of commonsense knowledge. 
Experiments show consistent improvement in models' performances.
\citet{CANDLE} further show that, by instantiating conceptualizations in abstract knowledge back to other novel instances, models can be further improved by training with newly instantiated knowledge.
\citet{DimonGen} also proposed a task that aims to generate diverse sentences describing concept relationships in various everyday scenarios.
Conceptualizations and associated abstract knowledge can further boost models' performances on this task.

\paragraph{Commonsense Question Answering:}
The task of commonsense question answering aims to answer questions that require commonsense knowledge. 
Various benchmarks and datasets have been proposed to evaluate LMs' performances, such as Abductive NLI (aNLI;~\cite{DBLP:conf/iclr/BhagavatulaBMSH20}), CommonsenseQA (CSQA;~\cite{DBLP:conf/naacl/TalmorHLB19}), PhysicalIQA (PIQA;~\cite{DBLP:conf/aaai/BiskZLGC20}), SocialIQA (SIQA;~\cite{DBLP:conf/emnlp/SapRCBC19}), and WinoGrande (WG;~\cite{DBLP:journals/cacm/SakaguchiBBC21}). 
To obtain a generalizable model for commonsense question answering, the most effective pipeline fine-tunes language models on QA pairs synthesized from knowledge in commonsense knowledge bases~\cite{DBLP:conf/aaai/MaIFBNO21,DBLP:conf/emnlp/ShiWFX0LS23,CAR}. 
The head $h_o$ and relation $r$ of a $(h_o, r, t)$ triple are transformed into a question using natural language prompts, with the tail $t$ serving as the correct answer option. 
Distractors or negative examples are generated by randomly sampling tails from triples that do not share common keywords with the head. 
To leverage conceptualization into the QA synthesis process, \citet{CAR,DBLP:journals/corr/abs-2403-07398} have proposed two strategies: 
On the one hand, they improve distractor sampling by incorporating conceptualizations of head events into common words of the question, thereby enabling selection of more relevant distractors that improve the model's ability to discern correct answers from distractors.
On the other hand, abstract knowledge derived from head events are integrated into original synthesized QA pairs, akin to COMET, to enrich the training data with diverse information, thereby enhancing the model's generalization capability in commonsense question answering tasks.
Experimental results show that the proposed strategies significantly improve the performance of commonsense question answering with conceptualization.

\subsection{Complex and Factual Reasoning}
\label{sec:downstream-reasoning-complex}
Complex reasoning refers to the ability to solve intricate problems that necessitate multiple steps of reasoning, which involves reasoning upon intricate scenarios, which may encompass multiple entities, events, and relations.
\citet{DBLP:journals/corr/abs-2403-07398} proposed to synthesize complex queries based on commonsense knowledge triples from ATOMIC. 
Both human-defined rules and tails generated by large language models are utilized to generate these complex queries. 
The model is subsequently trained on these complex queries to enhance its capability to solve complex reasoning problems. 
In this context, conceptualizations of head events can be used as augmentations to generate more diverse and complex queries~\cite{DBLP:journals/pvldb/CuiXWSHW17}. 
This can assist the model in learning to solve more intricate problems. 
Simultaneously, conceptualizations of head events can also be used to generate more informative distractors. 
This can aid the model in learning to distinguish more effectively between correct answers and distractors. 

\citet{StepBackPrompt} also developed a prompting method to improve the performance of LLMs on general and factual QA tasks.
It involves instructing the model with a simple zero-shot prompt to consider each question abstractly by generating and probing relevant concepts, then using this knowledge in the prompt to generate the answer.
This simple prompting method has been shown to significantly improve the performance of large language models on general QA tasks, including MMLU (Physics and Chemistry)~\cite{DBLP:conf/iclr/HendrycksBBZMSS21}, TimeQA~\cite{DBLP:conf/nips/ChenWWW21}, StrategyQA~\cite{DBLP:journals/tacl/GevaKSKRB21}, and MuSiQue~\cite{DBLP:journals/tacl/TrivediBKS22}. 
This work is interesting as it demonstrates that a simple prompting method can significantly enhance the performance of LLMs on general QA tasks.

\subsection{Others}
\label{sec:downstream-reasoning-others}
Aside from those two types of tasks, the line of works focusing on ultra-fine entity~\cite{DBLP:conf/acl/LevyZCC18,DBLP:journals/tacl/LiYC22,DBLP:conf/acl/DaiZ23,DBLP:conf/acl/JiangHJWXT23,DBLP:conf/emnlp/LiBS23,DBLP:conf/emnlp/FengPM23,DBLP:conf/acl/DaiSW20,DBLP:conf/emnlp/LiuLXH0W21,DBLP:conf/acl/OnoeBMD20} and event typing~\cite{DBLP:conf/aaai/Zhou00023,DBLP:conf/aaai/PepeBBN22,DBLP:conf/conll/ChenZWR20} can also be benefited by conceptualization.
These tasks aim to type named entities, nominal nouns, and pronouns into a set of free-form phrases.
Conceptualizations can serve as a bridge between the surface form and the target type, which is crucial for these tasks.

\section{Future Directions and Conclusions}
\label{sec:future-directions}
Finally, we conclude our work by discussing two interesting future directions.

\subsection{Controllable Generation}
Firstly, we envision that conceptualization can assist controllable text generation~\cite{DBLP:conf/acl/FengWL23,DBLP:conf/acl/HuangLLLSL23,DBLP:journals/csur/ZhangSLZS24}.
In some formulations, the task requires the model to generate a brief piece of text that remains consistent within a specific context or scope~\cite{DBLP:conf/nips/MengLPC22}. 
Conceptualizations can be applied as additional supervision signals or constraints that guide the model to generate text whose conceptualizations align with those in the input theme, thereby enhancing the controllability of the generated text. 
This could be achieved by training a pair of conceptualization generator and discriminator, which could be used to generate the conceptualizations and evaluate their consistency between input and output text. 
Conceptualization can also serve as data augmentation tools to provide more training data, preferably guided with human annotation or large language models as loose teachers, for training more robust text generators that better align with the controllable targeting data.

Similarly, it may also benefit hallucination reduction~\cite{DBLP:conf/acl/ChoubeyFVWLR23,DBLP:conf/acl/DaleVBC23,DBLP:conf/acl/JiLLYWZF23,DBLP:conf/acl/SunLMBL023}. 
Hallucination~\cite{DBLP:journals/csur/JiLFYSXIBMF23} refers to generating text that is unsupported by the input context, such as introducing information that is not present in the context or even contradicts it. 
In many reasoning scenarios, hallucination can be detrimental to the model’s performance, and neutralizing it is crucial for ensuring the reliability of the generated text. 
Towards this objective, conceptualization can be similarly applied as external signals to verify the generated text and ensure its accuracy. 
By measuring the semantic distance of conceptualizations between the given input and generated contents, hallucinations can possibly be detected by finding clearly unrelated concepts appearing at both ends. 
Empirical metrics to measure such distance can be the shortest path length of concepts in taxonomies such as WordNet~\cite{DBLP:journals/cacm/Miller95} and Probase~\cite{Probase}, or even embedding similarity between different concepts. 
However, it’s important to build a comprehensive set of conceptualizations of a given text to support such a verification process, as incomplete conceptualizations may cause erroneously detected hallucinations due to human-caused errors. 
We leave detailed implementations to future work.

\subsection{Modeling Changes in Distribution}
Conceptualization also plays a pivotal role in building reasoning systems that can capture situational changes in distribution to achieve System II reasoning~\cite{sloman1996empirical,kahneman2011thinking}.  
Among the several components that make up System II reasoning, a key element is the ability to reason with situational changes in distribution~\cite{DBLP:journals/cacm/BengioLH21,bengio2019system}.
These changes are triggered by environmental factors and actions by the agents themselves or others, especially when dealing with non-stationarities ~\cite{DBLP:journals/corr/abs-1709-08568,DBLP:journals/corr/abs-2505-07313,DBLP:journals/corr/abs-2505-16303}. 
This ability can be achieved by dynamically recombining existing concepts in the given environment or action and learning from the resultant situational changes~\cite{DBLP:conf/icml/LakeB18,DBLP:conf/iclr/BahdanauMNNVC19,DBLP:conf/nips/VriesBMCB19}. 
For instance, consider the event ``PersonX is driving a car on a sunny day.'' 
A change in the weather from sunny to rainy could cause a different outcome, such as ``PersonX becomes more cautious and drives slower.'' 
This illustrates that a change in weather conditions can lead to a change in the driver's behavior, representing an environmental change that triggers situational changes within the distribution of different weather conditions. 
In this process, the model is required to infer different changes that can possibly occur within a single event as the context, and reason about the potential outcome of each change. 
To model the distribution of different changes within an event, conceptualization can be used to represent the different states of the environment or action~\cite{MARS}. 
The model can then reason about the changes in distribution by manipulating the granularity of conceptualized changes. 
This type of distributional conceptualization not only provides an ontology for modeling the distribution of different changes within an event, but also assists the model in reasoning about the potential outcomes with appropriate abstract knowledge.
Future works can leverage LLMs to curate benchmark datasets via sequential conceptualization generation and develop advanced systems for System II reasoning.

\subsection{Conclusions}
In conclusion, this work surveys conceptualizations by proposing a four-level hierarchical definition and reviewing representative works in acquiring, leveraging, and applying entity and event-level conceptualization to downstream reasoning tasks. 
We also propose several intriguing ideas related to conceptualizations that may inspire further research. 
We hope our work paves the way for more research works toward generalizable machine intelligence through conceptualization and fosters the development of more advanced systems that can capture, organize, and learn world knowledge through connection between concepts, much like humans do.

\section*{Limitations}
The main limitations of our survey are two-fold. 
First, due to the vast amount of literature on conceptualization and conceptual knowledge across various datasets, we only cover the most representative works that stand out for their exceptional value and uniqueness in our taxonomy. 
Most of the papers are sourced from ACL Anthology\footnote{\href{https://aclanthology.org/}{https://aclanthology.org/}}, ACM Digital Library\footnote{\href{https://dl.acm.org/}{https://dl.acm.org/}}, and proceedings of leading artificial intelligence and machine learning conferences. 
Consequently, it is possible that some other related works are not included, but we aim to cover them in future versions.
Second, our survey specifically focuses on entity and event level conceptualization, leaving document/paragraph level and system level conceptualization unaddressed. 
However, it is impossible to survey everything within one single submission.
Future research can expand the scope of our survey to include more types of conceptualizations and modalities, such as categorization in the vision modality~\cite{DBLP:journals/jmlr/ChenW04}.

\section*{Ethics Statement}
Our paper presents a comprehensive survey of conceptualization, with a specific focus on entity and event levels. 
All datasets and models reviewed in this survey are properly cited and are available under free-access licenses for research purposes. 
We did not conduct additional dataset curation or human annotation work. 
Therefore, to the best of our knowledge, this paper does not yield any ethical concerns.

\section*{Acknowledgments}
We thank the anonymous reviewers and the area chair for their constructive comments. 
The authors of this paper were supported by the ITSP Platform Research Project (ITS/189/23FP) from the ITC of Hong Kong, SAR, China, as well as the AoE (AoE/E-601/24-N), the RIF (R6021-20), and the GRF (16205322) from the RGC of Hong Kong, SAR, China.

\bibliography{custom}

\appendix

\begin{center}
    {\Large\textbf{Appendices}}
\end{center}

\section{Conceptualization Acquisition Methods}
\label{appendix:methods}
In this appendix, we elaborate further on different methods of acquiring conceptualization and provide detailed explanations of their weaknesses.

\subsection{Extraction Based Methods}
For methods that follow the concept extraction paradigm, \citet{DBLP:conf/cikm/WangOWWLPG16} proposed a framework to optimize both tasks simultaneously, leading to stronger performances even compared to supervised concept extraction methods. 
\citet{DBLP:journals/pvldb/ParameswaranGR10} also proposed a market-basket-based solution, which adapts statistical measures of support and confidence to design a concept extraction algorithm that achieved high precision in concept extraction. 
\citet{DBLP:conf/www/RajagopalCOK13} proposed a solution to extract concepts from commonsense text, which uncovers many novel pieces of knowledge that cannot be found in the original corpora.
\citet{DBLP:conf/emnlp/HovyKR09,DBLP:conf/cikm/KrishnanSZ017,DBLP:conf/eacl/Pasca09} similarly proposed their solutions for large-scale concept extraction for more efficient data mining.

While these methods have been successful in extracting concepts and relationships from text, they have several limitations. 
First, they are heavily dependent on the quality of the text and the predefined list of concepts. 
If the text is noisy or contains many irrelevant words, the performance of these methods can degrade significantly, and the resulting extracted concepts may also tend to be noisy. 
Second, it's important to note that these methods primarily rely on parsing or pattern matching techniques on text and do not capture semantic information from the text~\cite{DBLP:conf/acl/Li0ZS25}. 
This potentially makes extracted concepts represented as isolated entities without any context or relationships and could result in mis-extraction of concepts or relationships, especially when the text contains ambiguous or polysemous words. 
For example, the word ``bank'' can refer to a financial institution, a river bank, or a memory bank, and without proper context, it's difficult to determine the correct meaning of it, thus leading to incorrect concept extraction. 
A low-performance parser, if wrongly parsing these words, may also lead to noisy results. 
Lastly, these methods are not able to generalize well to unseen concepts or text patterns that are not present in the predefined list of concepts. 
This limits their applicability to new domains or tasks that require the extraction of novel concepts or relationships. 
For example, to extract concepts from medical or legal domain text, specific patterns or extraction rules need to be designed, which may not be present when extracting normal conversational text.

\subsection{Retrieval Based Methods}
\subsubsection{Semantic-Based Retrieval}
To perform semantic-based retrieval, \cite{DBLP:conf/mm/NatsevHTXY07} proposed several approaches for semantic concept-based query expansion and re-ranking in multimedia retrieval, achieving consistent performance improvement compared to text retrieval and multimodal retrieval baseline. 
\cite{STC,ODSTC} improved text understanding by using a probabilistic knowledge base based on concepts and developed a Bayesian inference mechanism to conceptualize words and short text. 
Experimental results show significant improvements on text clustering compared to purely statistical methods and methods that use existing knowledge bases. 
\cite{DBLP:conf/cikm/KoopmanZBSL12} proposed a corpus-driven approach, adapted from LSA, to retrieve medical concepts with semantic similarity measures. 
\cite{DBLP:conf/emnlp/ZhengY15} similarly used topic modeling and key concept retrieval methods to construct queries from electronic health records, which significantly improves the retrieval of tailored online consumer-oriented health education materials.

Although these methods have shown promising results in various domains, they have several limitations. 
First, the performance of semantic-based retrieval heavily relies on the quality of the knowledge base or concept taxonomy. 
In other words, it requires the knowledge base to be comprehensive, accurate, hierarchical, and up-to-date. There are very few knowledge bases that meet all these requirements, and constructing such a knowledge base is a non-trivial task. 
With incomplete knowledge bases, which are common in practice, the performance of semantic-based retrieval methods can be significantly degraded. 
Second, semantic-based retrieval methods are usually computationally expensive, as they require calculating the similarity between the input instance and all concepts in the knowledge base. 
This can induce exponentially increasing computational cost as the size of the knowledge base grows.
When dealing with large-scale applications, this even becomes infeasible. 
Though caching and indexing techniques can be used to speed up the retrieval process, they are not always effective and cannot generalize well when unseen concepts or instances are encountered. 
Third, semantic-based retrieval methods still do not consider the semantic context of the input instance. 
A straightforward formulation is that the model treats the input instance as a bag of words and ignores the word order and syntactic structure. 
This can lead to a loss of important semantic information, especially when the input instance is long and complex. 
In this case, the semantic similarity between the input instance and the concepts in the knowledge base may not reflect the true semantic relevance. 

\subsubsection{Neural-Based Retrieval}
For neural-based retrieval, aside from~\citet{AbstractATOMIC}, \cite{CourseConceptExtraction} similarly proposes a novel three-stage framework, which leverages the power of pre-trained language models explicitly and implicitly and employs discipline-embedding models with a self-train strategy based on label generation refinement across different domains.

To deal with the large amount of unlabeled data after human annotation, \cite{CAT} further proposed a semi-supervised method to unlabel the data with a supervised trained conceptualization discriminator. 
The discriminator is trained to rate the plausibility of unlabeled conceptualization and the model will be further refined by training on a concatenation of labeled and unlabeled data. 
This results in a significant improvement in the performance of the conceptualization discriminator, thus enhancing the quality of the retrieved concepts.

Despite these promising results in concept retrieval, neural-based retrieval methods have several limitations. 
First, these methods are usually data-hungry and require a large amount of labeled data for training. 
This can be a bottleneck in practice, as labeling data is often expensive and time-consuming. 
Human annotations are usually required to collect such data, and for models to be generalizable across different domains, the labeled data should be diverse and representative. 
This is even more costly and challenging to obtain. Second, neural-based retrieval methods still rely on the coverage and quality of the knowledge base or concept taxonomy. 
If the knowledge base is incomplete or inaccurate, the performance of neural-based retrieval methods can be significantly affected. 
Moreover, they cannot generate new concepts or instances that are not in the knowledge base, which limits their generalization ability. 

\subsection{Generative-Based Methods}
\subsubsection{Fine-Tuning-Based Generative Methods}
While most fine-tuning based methods are explicitly discussed in the main body, we explain their limitations here. 
First, these methods are usually computationally expensive, as they require fine-tuning a large pre-trained language model on a specific dataset. 
Both the fine-tuning and the training data collection process can be time-consuming and resource-intensive. 
Extensive crowd-sourcing or human annotations are usually required to collect high-quality training data, which can be costly and challenging to obtain when the domain coverage scales up. 
Second, the feasibility of fine-tuning-based generative methods on other domains, such as medical or legal text, is still an open question. 
The performance of these methods heavily relies on the quality and diversity of the training data, and it's not clear how well they can generalize to new domains or tasks as text understanding abilities vary across different domains. 
For social commonsense, pre-trained language models have shown strong performance possibly due to a large overlap in the training data distribution, but for other domains, the performance is still unclear. 
Lastly, although existing studies have shown that fine-tuning based generators can deliver novel concepts that are not in the training data, such a ratio is relatively low and the quality of the generated concepts is still not as good as human annotated ones. 
This is expected as the models are fitted into the distribution of the training data, and it's hard for them to generate concepts that are out of the distribution. 

\subsubsection{Zero-Shot Generative Methods}
Zero-shot generative methods aim to generate the desired output for any task's input without any task-specific fine-tuning. 
A very representative example of such generative models is the recently popularized LLMs~\cite{openai2022chatgpt,GPT4,llama,llama2,Gemma,Gemini}. 
These models have been pre-trained on very large corpora, including those from the web, Wikipedia, books, and more, and have shown strong performance in various natural language processing tasks, including text generation~\cite{DBLP:conf/acl/MaynezAG23,DBLP:conf/aaai/ChenXWLM24, DBLP:conf/acl/WangCLNX0SLGLYB25}, temporal reasoning~\cite{DBLP:conf/acl/TanNB23,DBLP:conf/www/YuanXHA24}, causal reasoning~\cite{DBLP:conf/eacl/ChanCWJFLS24,DBLP:conf/eacl/DalalBA23,DBLP:conf/nips/JinCLGKLBAKSS23}, commonsense reasoning~\cite{DBLP:conf/emnlp/JainSA0JD23,DBLP:journals/corr/abs-2303-16421,DBLP:conf/www/FangZWSH21,DBLP:conf/emnlp/FangWCHZSH21,DBLP:conf/emnlp/DengW0LS23}, logical reasoning~\cite{DBLP:conf/acl/WangFYSWS23,DBLP:conf/iclr/WangSWS23,DBLP:conf/nips/WangYS21,DBLP:conf/nips/BaiLW0S23}, and more~\cite{DBLP:conf/emnlp/QinZ0CYY23,DBLP:conf/emnlp/ChengQCFWCRGZSZ23,DBLP:journals/corr/abs-2404-13627, DBLP:journals/corr/abs-2504-10284}.

In the context of conceptualization acquisition, zero-shot generative methods aim to generate conceptualizations for instances without any instance-conceptualization pairs in the training data. 
\citet{CANDLE} proposed a few-shot knowledge distillation method to distill conceptualizations and associated abstract inferential knowledge from a large language model to a large-scale knowledge base. 
\citet{AbsPyramid} also proposed acquiring conceptualizations for entities and events in ASER by instructing ChatGPT with a few-shot prompt. 
They further designed an instruction-tuning based method to evoke more conceptualizations from large language models by fine-tuning them with explanations on how the conceptualization is derived from the instance and their plausible reasoning chains~\cite{AbsInstruct}.
\citet{StepBackPrompt} proposed a simple prompting technique, inspired by chain-of-thought reasoning, that enables LLMs to do conceptualizations to derive high-level concepts and first principles from instances containing specific details. 
\citet{HIConcept} advanced this idea by proposing to extract predictive high-level features (concepts) from a large language model’s hidden layer activations.

The benefits of these methods are twofold. 
First, such generation can introduce conceptualizations at a very low cost, as the models are pre-trained and do not require any task-specific fine-tuning. 
The only burden seems to be deployment and inference cost, which require a large amount of computational resources and time for large-scale generation. 
However, compared to all previous fine-tuning-based methods, zero-shot generative methods are much more efficient and scalable, as they do not require any training data or fine-tuning process.
Second, zero-shot generative methods have shown strong generalization capabilities to new instances and domains. 
They can generate conceptualizations for instances that are not in the training data and have shown strong performance in various conceptualization acquisition tasks. 
This is particularly useful when the training data is scarce or when the domain is new, and there are no existing training data available. 
Since these large language models are pre-injected with vast amounts of knowledge, this makes generalization possible.

\end{document}